\documentclass[conference]{IEEEtran}
\IEEEoverridecommandlockouts
\usepackage{cite}
\usepackage{amsmath,amssymb,amsfonts}
\usepackage{algorithmic}
\usepackage{graphicx}
\usepackage{textcomp}
\usepackage{xcolor}

\usepackage{mathptmx} 
\usepackage{times} 
\usepackage{amsmath} 
\usepackage{amssymb}  
\usepackage{array}
\usepackage{caption}
\usepackage{subcaption}
\usepackage{balance}
\usepackage{float}
\usepackage{url}
\usepackage{multirow}
\usepackage[export]{adjustbox}
\usepackage{footnote}
\usepackage{booktabs}
\usepackage{multirow}
\usepackage{enumitem}
\usepackage{soul}
\setlist{nosep, leftmargin=14pt}
\usepackage{makecell}

\def\BibTeX{{\rm B\kern-.05em{\sc i\kern-.025em b}\kern-.08em
    T\kern-.1667em\lower.7ex\hbox{E}\kern-.125emX}}
\begin{document}

\title{Slice-level Detection of Intracranial Hemorrhage on CT
Using Deep Descriptors of Adjacent Slices
}

\author{\IEEEauthorblockN{Dat T. Ngo}
\IEEEauthorblockA{
\textit{Vingroup Big Data Institute}\\
\textit{10000, Hanoi, Vietnam} }
\and
\IEEEauthorblockN{Thao T.B. Nguyen}
\IEEEauthorblockA{\textit{Vingroup Big Data Institute}\\
\textit{10000, Hanoi, Vietnam} }
\and
\IEEEauthorblockN{Hieu T. Nguyen}
\IEEEauthorblockA{\textit{Vingroup Big Data Institute}\\
\textit{10000, Hanoi, Vietnam} }
\and
\IEEEauthorblockN{Dung B. Nguyen}
\IEEEauthorblockA{\textit{Vingroup Big Data Institute}\\
\textit{10000, Hanoi, Vietnam} }
\and
\IEEEauthorblockN{ \hspace*{3cm} Ha Q. Nguyen}
\IEEEauthorblockA{\hspace*{3cm} \textit{Vingroup Big Data Institute}\\
\textit{ \hspace*{3cm} VinBigData JSC,} \\
\hspace*{3cm} \textit{10000, Hanoi, Vietnam} }
\and
\IEEEauthorblockN{Hieu H. Pham}
\IEEEauthorblockA{\textit{VinUni-Illinois Smart Health Center,}\\
\textit{College of Engineering \& Computer Science,} \\
\textit{10000, Hanoi, Vietnam} \\ Correspondence: \textcolor{blue}{hieu.ph@vinuni.edu.vn} }

}

\maketitle

\begin{abstract}
The rapid development in representation learning techniques such as deep neural networks and the availability of large-scale, well annotated medical imaging datasets have to a rapid increase in the use of supervised machine  learning in the 3D medical image analysis and diagnosis. In particular, deep convolutional neural networks (D-CNNs) have been key players and were adopted by the medical imaging community to assist clinicians and medical experts in disease diagnosis and treatment. However, training and inferencing deep neural networks such as D-CNN on high-resolution 3D volumes of Computed Tomography (CT) scans for diagnostic tasks poses formidable computational challenges. This challenge raises the need of developing deep learning-based approaches that are robust in learning representations in 2D images, instead 3D scans. In this work, we propose for the first time a new strategy to train \emph{slice-level} classifiers on CT scans based on the descriptors of the adjacent slices along the axis. In particular, each of which is extracted through a convolutional neural network (CNN). This method is applicable to CT datasets with per-slice labels such as the RSNA Intracranial Hemorrhage (ICH) dataset, which aims to predict the presence of ICH and classify it into 5 different sub-types. We obtain a single model in the top 4\% best-performing solutions of the RSNA ICH challenge, where model ensembles are allowed. Experiments also show that the proposed method significantly outperforms the baseline model on CQ500. The proposed method is general and can be applied for other 3D medical diagnosis tasks such as MRI imaging. To encourage new advances in the field, we will make our codes and pre-trained model available upon acceptance of the paper.
\end{abstract}

\begin{IEEEkeywords}
component, formatting, style, styling, insert
\end{IEEEkeywords}

\section{Introduction}

\begin{figure}[ht]
  \centering
  \includegraphics[width=8.5cm,height=5.5cm]{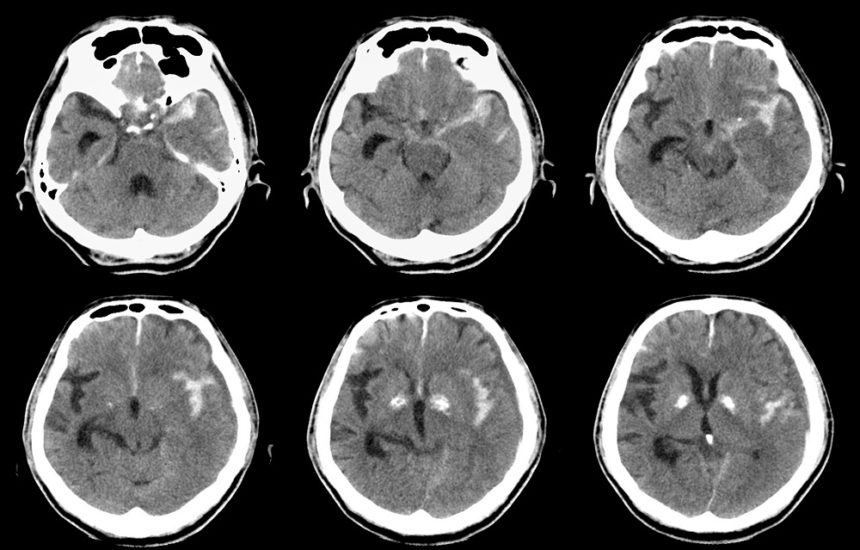}
 \caption{\normalsize{Examples of head computed tomography scans. Each study may contain from hundreds to thousands slices, which makes training deep neural networks on this modality challenging in terms of computing requirements.}}
  \label{fig:1}
\end{figure}

Deep neural networks, in particular deep convolutional neural networks (CNNs) \cite{yamashita2018convolutional,alakwaa2017lung,dao2022phase}  became a key tool in medical imaging analysis 
\cite{serte2021deep,singh20203d,zheng2020deep,gruetzemacher20183d}. In particular, deep neural networks have showed excellent performance in 2D medical imaging data such as the interpretation of X-ray scans \cite{rajpurkar2017chexnet,shen2019deep,pham2021interpreting,pham2021accurate,nguyen2021clinical,nguyen2021novel,tran2021learning,nguyen2021vindr}. However, training very deep networks on high-resolution 3D volumes of Computed Tomography (CT) scans requires a huge computing resource. For examples, each typical 3D CT scans study may contain from hundreds to thousands slices, which makes training deep neural networks on this modality challenging in terms of computing requirements (see Figure \ref{fig:1}). In this study, we aim to develop an efficient 2D deep learning-based approach for 3D medical imaging analysis  that is able to provide the same level of performance compared to 3D-based learning approaches. To tackle this challenge, we consider each 3D CT scan as a set of 2D images and propose a new strategy to train \emph{slice-level} classifiers on CT based on the descriptors of the adjacent slices along the axis, each of which is extracted through a convolutional neural network (CNN). Extracted features are then concatenated to produce better representations for classification tasks. This method is applicable to CT datasets with per-slice labels such as the RSNA Intracranial Hemorrhage (ICH) dataset~\cite{RSNA}, which aims to predict the presence of ICH and classify it into 5 different sub-types. 

The proposed method exploits a two-stage training scheme. In the first stage, we treat a CT scan simply as a set of 2D images and train a state-of-the-art CNN classifier~\cite{He-etal:2016} that was pretrained on ImageNet~\cite{krizhevsky2012imagenet}. During the training process, each slice is sampled \emph{together} with the 3 slices before and the 3 slices after it. By this way, each CNN model takes a batch size a multiple of 7 slices as input. In the second stage, the output \emph{descriptors} of each block of 7 consecutive slices obtained from the first stage are stacked into an image and fed to another CNN model for final prediction of the middle slice. Then, we leverage the assemble learning to boost classification performance. To evaluate the effectiveness of the proposed approach, we train our deep learning framework on the RSNA dataset and additionally evaluated on the CQ500 dataset~\cite{Chilamkurthy-etal:2018}, which adopts the same a set of labels but only on study level. We obtain a single model\footnote{Model weights and codes are available upon acceptance of the paper.} in the top 4\% best-performing solutions of  the RSNA ICH challenge, where model ensembles are allowed. Experiments also show that the proposed method significantly outperforms the baseline model on CQ500~\cite{Chilamkurthy-etal:2018}. To summarize, our key contribution is to introduce a simple 2D-based technique that allows to train a very deep neural network on 3D imaging data and providing better results compared to 3D-based approaches. The proposed method is simple, general can can be applied for many other 3D medical imaging analysis tasks such as MRI imaging analysis. The rest of the paper is organized as follows. The proposed approach is presented in Section \ref{sect:2}. Section \ref{sect:3} describe the experiments and discusses the experimental results. Finally, section \ref{sect:4} concludes the work as well as presents its perspectives.\\[-0.5cm]
\section{Proposed approach}
\label{sect:2} 
\begin{figure*}[ht]
  \centering
  \includegraphics[width=18cm,height=7.5cm]{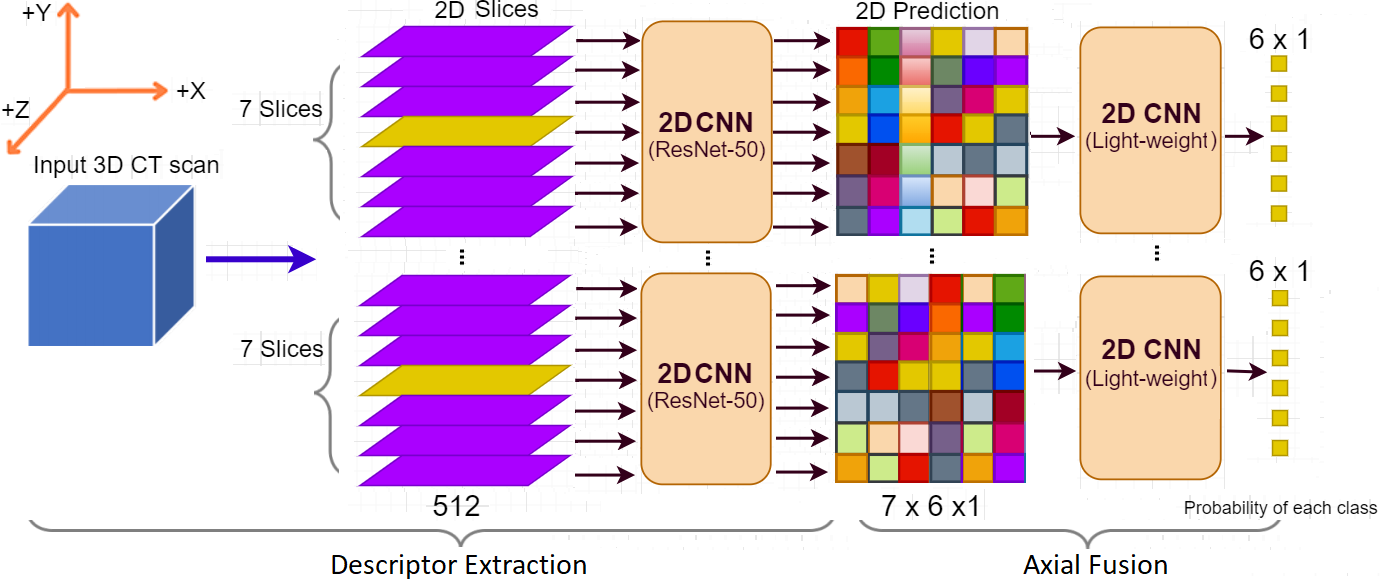}
 \caption{\normalsize{Illustration of the proposed two-stage training procedure.}}
  \label{fig:2}
\end{figure*}

Non-contrast head CT scans are the current standard imaging propotcol for initial imaging of patients with stroke symptoms, including intracranial haemorrhag \cite{chilamkurthy2018deep}. We aim in this study to develop and validate a 2D-based deep learning algorithms for automated detection of the key findings from head CT scan scans called intracranial haemorrhage. The proposed approach is validated on the RSNA Intracranial Hemorrhage (ICH) dataset. It is difficult to exploit high-performing deep neural networks for the ICH classification while keeping the full 3D resolution of the input CT. On the other hand, using slices as independent 2D images and ignoring the axial information of data causes detrimental effects to the algorithm's performance. Our method gets the best of both worlds: we apply transfer learning~\cite{pan2009survey} on 2D images to perform classification per slice and then assemble the results of local slices to refine the prediction of the middle one. The proposed scheme, illustrated in Figure~\ref{fig:2}, consists of two stages: descriptor extraction and axial fusion. Below we describe these element in details. 

\subsection{Descriptor extraction}
In this stage, we train a 2D CNN to classify individual slices of the CT scans, which are converted into 3-channel images using 3 different windows: brain ($l=40, w=80$), subdural ($l=75, w=215$), and bone ($l=600, w=2800$). The output of the network is a descriptor of size 6$\times$1 that includes the probabilities of the 5 ICH sub-types and an additional class for any of them. During training, each slice is always sampled in a block of 7 that includes itself in the center and 6 neighboring slices. With this approach, we can take advantage of pre-trained models on ImageNet~\cite{Deng-etal:2009} to initialize the network. Specifically, a ResNet-50~\cite{He-etal:2016} was used in our experiments. We followed the procedure in~\cite{He-etal:2019} and trained the network for 20 epochs with a batch size of $16\times 7$ using Adam optimizer~\cite{kingma2014adam}. An initial learning rate of $5e-4$ and the cosine annealing learning rate scheduler~\cite{loshchilov2016sgdr} were used. Several augmentation techniques such as cropping, resizing, flips, rotations, distortions, gaussian noise, and CutMix~\cite{Yun-etal:2019} were applied to prevent the network from overfitting.  
\subsection{Axial fusion}
This stage combines the descriptors of each 7 consecutive slices generated in stage 1 to exploit the axial information and to refine the prediction of the centered slice. In particular, we concatenate the 7 descriptors into a 7$\times$6$\times$1 tensor and train a 3-layer CNN to output the final classification result for the representative slice in the middle of the block. This network contains only 2 convolution layers and 1 fully connected layer. The 2D convolution kernels help the model learn both the relationship between ICH predictions across local slices and the relationship between probabilities for the sub-types. The output of the fusion network can therefore be seen as a re-calibrated prediction of a single slice.

\section{Experiments and results}
\label{sect:3} 
\subsection{Datasets and evaluation protocols}

The RSNA and CQ500 datasets were used to verify the effectiveness of the proposed approach. Both of them contain non-contrast CT scans that are labeled with 5 sub-types of ICH: \textit{intraparenchymal}, \textit{intraventricular}, \textit{subdural}, \textit{extradural}, and \textit{subarachnoid}. The only difference between the two datasets is that the labels of RSNA are per slice, while the those of CQ500 are per CT scan. The whole RSNA dataset was split into 3 parts: a public training set (19,530 studies), a public testing set (2,214 studies), and a private testing set (3,518 studies). Each study is a CT scan of 20 to 60 slices of 512$\times$512 pixels. The weighted log loss was used as the evaluation metric for this dataset, in which a weight of $2/7$ was used for the ICH label and a weight of $1/7$ was used for each of the 5 sub-types. 

Meanwhile, CQ500 consists of 500 studies, from which 490  were selected for experiments while the rest 10 of them are noisy and were excluded from the dataset. We used CQ500 as another test set to validate the efficiency and robustness of the proposed algorithm, which was merely trained on the public training set of RSNA. On this test, the performance of our method is measured by area under the ROC curve (AUC). Note that the study-level probability of any class was taken as the maximum probability amongst all slices.\\[-0.5cm]
\subsection{Experimental results}

We report a weighted log loss of 0.05341 on the private testing set of RSNA, which ranks in top 4$\%$ over 1345 teams on the Kaggle leaderboard. Note that our result is provided by a single ResNet-50 model, while many other solutions in this competition exploit ensemble techniques. On CQ500, the proposed method achieves a mean AUC of 0.971. This is an improvement of around $2\%$ compared to the baseline model~\cite{Chilamkurthy-etal:2018}. Especially, our method provides better AUC scores over all disease labels as shown in Table~\ref{table:1}. These results strongly demonstrate the generalization capacity of our model, which was trained on a different dataset with a different labeling protocol.
\begin{table}[h!]
\begin{center}
\begin{tabular}{l c c c} 
 \hline
\normalsize{\textbf{Findings}} &  \normalsize{\cite{Chilamkurthy-etal:2018}} &  & \normalsize{\textbf{Ours}}  \\
\hline
\hline
 \normalsize{ICH (any subtypes)} &  \normalsize{0.9419} &  &  \normalsize{\textbf{0.9612}} \\
 \normalsize{Intraparenchymal} &  \normalsize{0.9544} &  &  \normalsize{\textbf{0.9691}} \\
 \normalsize{Intraventricular} &  \normalsize{0.9310} &  &  \normalsize{\textbf{0.9832}} \\
 \normalsize{Subarachnoid} &  \normalsize{0.9574} &  &  \normalsize{\textbf{0.9596}} \\
\normalsize{Subdural} &  \normalsize{0.9521} &  &  \normalsize{\textbf{0.9694}} \\ 
\normalsize{Extradural}  &  \normalsize{0.9731} &  &  \normalsize{\textbf{0.9814}} \\
 \hline
 \hline
\normalsize{Mean}  &  \normalsize{0.9520} &  &  \normalsize{\textbf{0.9710}} \\
 \hline
\end{tabular}
\caption{Experimental results measured by AUC score on CQ500 dataset.\\[-1cm]}
\label{table:1}
\end{center}
\end{table}
\\[-1cm]
\section{Conclusion}
\label{sect:4} 
Training very deep neural networks such as D-CNN model on 3D medical scans like CT or MRI is a challenging task. Three-dimensional computed tomography (3DCT) is a type of CT scanning which records multiple images over time, that requires huge computational requirements when D-CNNs are explored.  We presented in this paper a novel two-stage training strategy for the task of slice-level classification on CT scans. The key idea is to sample each slice together with its neighbors and to refine the classification of it using the coarse descriptors of the whole group. This method was experimentally demonstrated to work well on ICH datasets like RSNA and CQ500. We believe that it can be easily extended to other 3D datasets of CT or MRI scans.\\
\\
\noindent \textbf{Acknowledgements}.
This study was supported by VinBigData JSC.

\bibliographystyle{abbrv}

\bibliography{refs}
\end{document}